%% file: icdm_main.tex
\documentclass[conference]{IEEEtran}
\IEEEoverridecommandlockouts

\usepackage{cite}
\usepackage{amsmath,amssymb,amsfonts}
\usepackage{algorithmic}
\usepackage{graphicx}
\usepackage{textcomp}
\usepackage{xcolor}

\input{constants}
\usepackage{hyperref}
\usepackage{comment}
\usepackage{multirow}
\usepackage{subcaption}
\usepackage{booktabs}
\usepackage{array}
\usepackage{csquotes}

\newtheorem{example}{Example}
\newtheorem{theorem}{Theorem}
\newtheorem{lemma}{Lemma}
\newtheorem{definition}{Definition}

\newtheorem{proposition}{Proposition}
\newtheorem{property}{Property}
\newcommand{\reconstructionError}{\epsilon}
\newcommand{\featureWeights}[1]{\omega_{\mathbf{x}}^{#1}}
\newcolumntype{C}[1]{>{\centering\arraybackslash}p{#1}}

\def\BibTeX{{\rm B\kern-.05em{\sc i\kern-.025em b}\kern-.08em
    T\kern-.1667em\lower.7ex\hbox{E}\kern-.125emX}}
\begin{document}

\title{\paperTitle
}

\author{\IEEEauthorblockN{1\textsuperscript{st} Li Rong Wang}
\IEEEauthorblockA{
\textit{College of Computing and Data Science (CCDS)} \\
\textit{Nanyang Technological University (NTU)}\\
\textit{A*STAR Centre for Frontier AI Research}\\
Singapore \\
}
\and
\IEEEauthorblockN{2\textsuperscript{nd} Jamie Duell}
\IEEEauthorblockA{\textit{School Of Computing and Digital Technologies} \\
\textit{Sheffield Hallam University}\\
South Yorkshire, England \\
}
\and
\IEEEauthorblockN{3\textsuperscript{rd} Xinran Xu}
\IEEEauthorblockA{\textit{CCDS} \\
\textit{NTU}\\
Singapore \\
}
\and
\IEEEauthorblockN{4\textsuperscript{th} Thomas C. Henderson}
\IEEEauthorblockA{\textit{School of Computing} \\
\textit{University of Utah}\\
United States \\
}
\and
\IEEEauthorblockN{5\textsuperscript{th} Yu Yue Hew}
\IEEEauthorblockA{\textit{Department of Haematology} \\
\textit{Tan Tock Seng Hospital (TTSH)}\\
Singapore \\
}
\and
\IEEEauthorblockN{6\textsuperscript{th} Pik Wan Erica Chiang}
\IEEEauthorblockA{\textit{Department of Haematology} \\
\textit{TTSH}\\
Singapore \\
}
\and
\IEEEauthorblockN{7\textsuperscript{th} Xiao Wei Al'star Ang}
\IEEEauthorblockA{\textit{Department of Haematology} \\
\textit{TTSH}\\
Singapore \\
}
\and
\IEEEauthorblockN{8\textsuperscript{th} Bingwen Eugene Fan}
\IEEEauthorblockA{\textit{Lee Kong Chian School of Medicine} \\
\textit{NTU}\\
Singapore \\
}
\and
\IEEEauthorblockN{9\textsuperscript{th} Xiuyi Fan}
\IEEEauthorblockA{\textit{Lee Kong Chian School of Medicine, CCDS} \\
\textit{NTU}\\
Singapore \\
}
}

\maketitle

\begin{abstract}
    Artificial intelligence has strong potential to support clinical decision-making, yet its adoption in healthcare remains limited due to a lack of trust. 
    Uncertainty estimation can signal unreliable predictions, and explainable AI (XAI) can clarify how predictions are made but existing methods treat them separately, providing no feature-level insight into why a prediction is uncertain or which tests to prioritize to reduce it.
    To address this gap, we propose explainable uncertainty estimation, which unifies uncertainty estimation and XAI to both quantify uncertainty and explain feature-level contributions. 
    We introduce the Expected Gradients Reconstruction Uncertainty Estimate (egRUE), which incorporates prediction explanations into its uncertainty computation and decomposes uncertainty into feature-wise contributions. 
    We prove theoretical properties of egRUE and show through experiments that it improves reliability and interpretability compared to existing methods.
    A user study with medical experts further demonstrates that egRUE's explanations improve calibrated trust over uncertainty scores alone, increasing confidence in correct predictions and reducing confidence in incorrect ones. 
    By combining prediction uncertainty with feature-level explanations, egRUE strengthens decision-making support in safety-critical healthcare settings, clarifying both when predictions may be unreliable and which features drive that uncertainty.
\end{abstract}

\begin{IEEEkeywords}
Uncertainty Estimation, Uncertainty Attribution, Explainability
\end{IEEEkeywords}

\section{Introduction}
Artificial intelligence shows promise for clinical decision-making \cite{sheliemina2024use,rawat2025introduction} but its deployment remains constrained by a lack of trust \cite{alonso2025ai,hassan2024barriers}. Conventional models give definitive predictions, even though uncertainty is intrinsic to medicine. As William Osler famously noted, \emph{\enquote{Medicine is a science of uncertainty and an art of probability.}} Uncertainty estimation in Machine Learning (ML) embraces this inherent uncertainty by quantifying how uncertain a model is about its predictions, thereby enhancing model trustworthiness. It helps users trust the model when it is confident and exercise caution when the model is uncertain and likely to make errors \cite{marusichUsingAIUncertainty2024,toh2025effect}. Notable methods include Gaussian process regressors  \cite{williams1995gaussian}, Monte Carlo dropout \cite{galDropoutBayesianApproximation2016} and deterministic uncertainty quantification \cite{van2020uncertainty}. They produce uncertainty estimations but do not attribute uncertainty to specific sources. This limitation has been noted in attempts to connect uncertainty with explanation \cite{watson2023explaining,wangOptimizationBasedUncertaintyAttribution2025,NEURIPS2023_9f0b1220}.

Explainable Artificial Intelligence (XAI) \cite{yang2023survey} increases the trustworthiness of ML models by explaining how the model derives its predictions. Explanations are presented as feature-attribution values -- values specific to each feature in the input that quantify its importance to the prediction. Popular methods include Integrated Gradients (IG) \cite{IG} and SHapley Additive exPlanations (SHAP) \cite{Lundberg}. They explain predictions but do not address uncertainty.

In their seminal work on understanding uncertainty in decision-making within cognitive science and human behavior study, \cite{lipshitz1997coping} propose that: 
\begin{quote} 
{\em \enquote{Different types of uncertainty can be classified based on their issues (i.e., what the decision maker is uncertain about) and their sources (i.e., what causes this uncertainty).}} 
\end{quote}
and stated how better addressing of the {\em source} of uncertainty can eventually lead to improved decision-making. A computerised example of their theory in ML is the following.

\vspace{5pt}
\begin{example}
\label{exp:open}
Consider a lung cancer screening model with a subject having characteristics in the {\bf Value} columns. The prediction question is whether the subject has lung cancer. 
\vspace{3pt}
\begin{small}
\begin{center}
\textup{
\begin{tabular}{|c|c|c||c|c|c|}
\hline
   {\bf Feature} & {\bf Value} & {\bf U-Exp} & {\bf Feature} & {\bf Value} & {\bf U-Exp} \\
\hline
\hline
   Alcohol & Yes  & 0.11 & Smoker & Yes & 0.48 \\
   Sex     & Male & 0.24 & Age    & 71  & 0.07 \\
   BMI     & 17.7 & 0.00 & Diet Score & 52 & 0.00\\
   SBP     & 126  & 0.00 & DBP        & 70 & 0.00 \\
   \hline
   \multicolumn{6}{|c|}{(\ldots another 11 features with different U-Exps\ldots)}\\
   \hline
   \hline
   \multicolumn{3}{|l|}{{\bf Prediction:} Positive} &
   \multicolumn{3}{|l|}{{\bf Overall Uncertainty:} 0.916}\\   
\hline     
\end{tabular}
}
\end{center}
\end{small}
\vspace{7pt}

This is a difficult prediction because, although the aging male drinks and smokes, he is otherwise healthy with a good diet and BMI value. The prediction model gives an incorrect {\em positive} prediction. However, our explainable uncertainty estimation algorithm attributes the high uncertainty (0.48 in the {\bf U-Exp} column) to the feature {\em Smoker}. 

An oncologist interpreted this with:
\begin{quote}
The binary feature \enquote{Smoker} is insufficient. We must know how much he smokes.
\end{quote}
\end{example}

From Example~\ref{exp:open}, we can see that knowing a model is uncertain about its predictions and understanding how a prediction is computed brings more insight to underlying problem. 
This is pivotal, as the use of XAI without uncertainty quantification can lead to misguided trust in model predictions and assigned importance \cite{make6020055}. Thus, it is critical to combine uncertainty estimation and XAI to facilitate interpretable uncertainty quantification and leverage explanations in uncertainty calculations.

Several methods have recently attempted this combination,  notably generative counterfactual explanations like CLUE~\cite{antoran2020getting}. While CLUE synthesizes \enquote{certain} inputs to explore uncertainty, our work instead uses \textit{Feature Attribution} to identify specific morphological markers driving uncertainty in the original clinical image. Unlike CLUE, we avoid auxiliary generative models, ensuring faithfulness to patient data.

In this context, the contributions of this work are:
\begin{enumerate}
    \item Quantify and explain uncertainty in ML predictions;
    \item Provide a theoretical evaluation of the proposed method; 
    \item Evaluate our methods with practicing clincians. 
\end{enumerate}    

\begin{figure}
    \centering
    \includegraphics[trim={0 10.8204cm 18.034cm 0},clip,width=0.9\linewidth]{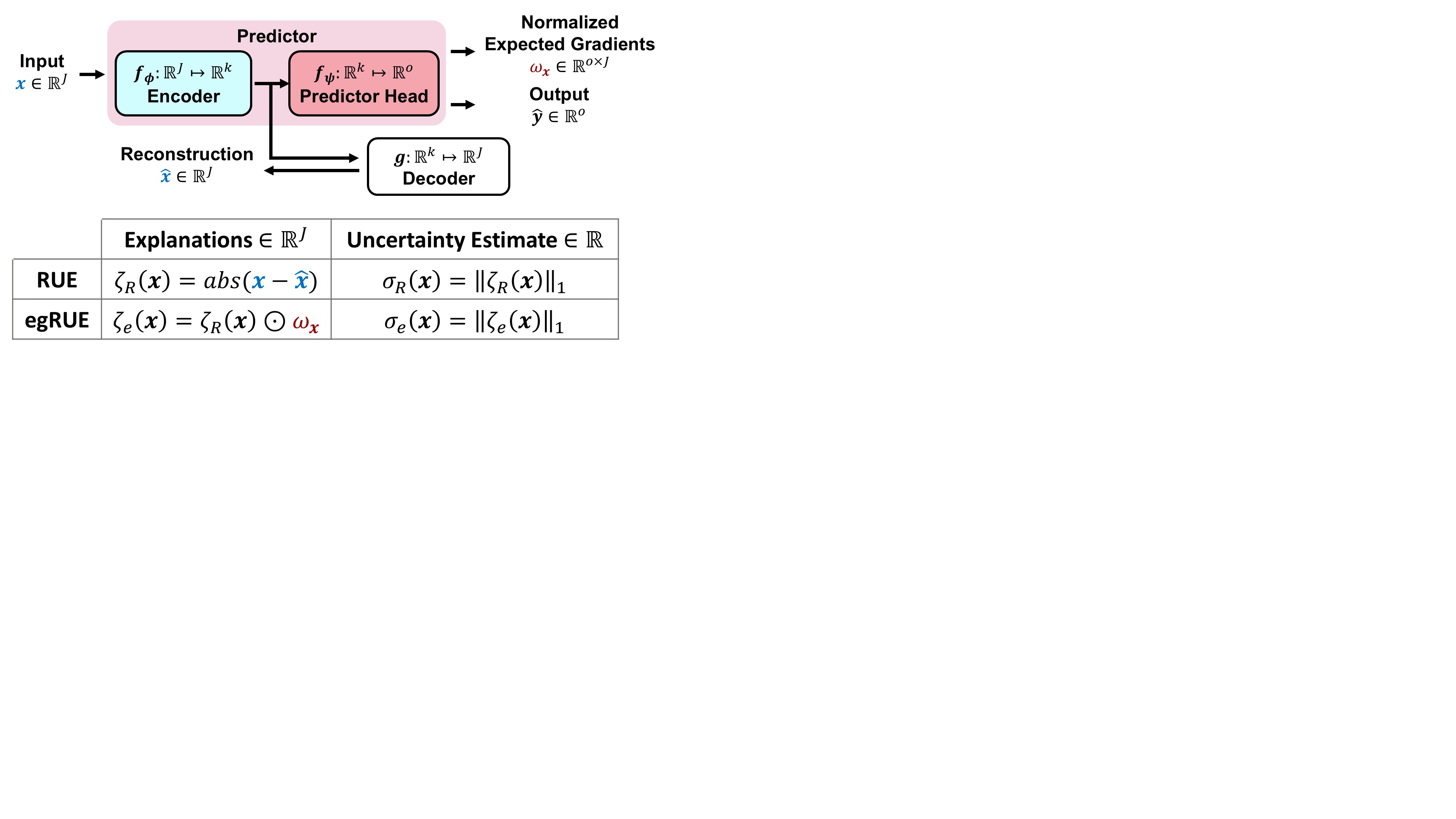}
    \caption{System architecture for RUE and egRUE.}
    \label{fig:architecture}
    \vspace{-15pt}
\end{figure}

For the remainder of this paper, after a very brief introduction to the XAI technique used in this work, we introduce the Expected Gradients Reconstruction Uncertainty Estimate (egRUE) (Figure~\ref{fig:architecture}). We present theoretical results and provide quantitative experiments on four real-world datasets, demonstrating the superiority of egRUE. To validate the practical utility of our uncertainty explanations, we further conduct a user study with medical experts, showing that egRUE's explanations improve calibrated trust over uncertainty scores alone. Unlike most existing works, which often overlook explaining their estimations, our approach is unique in offering uncertainty explanations within a single prediction model.

\section{Preliminary}
The key XAI technique used in this work is the {\em Expected Gradients (EG)} \cite{Erion2021}, which is a {\em feature-attribution} method. We briefly introduce them first.

    Given a function $f : \mathbb{R}^{J} \rightarrow \mathbb{R}$, and an instance $\mathbf{x} = \langle x^1, \ldots, x^J \rangle$, a {\em feature-attribution method} $\phi : \mathbb{R}^J \rightarrow \mathbb{R}^J$ produces a $J$ dimensional vector of {\em feature-attribution values},
    such that $\phi(\mathbf{x}) = \langle \phi^1, \ldots, \phi^J \rangle$ represents each feature's contribution towards the function's output. 

Given a prediction function $f$, EG unveils the impact of each input dimension of $f$ to its output by computing feature attribution values. Specifically, EG averages the gradients of the prediction function $f$ over a distribution of baseline inputs, in which the gradient of the output with respect to the input $\mathbf{x}$ is computed with:
\begin{equation*}
    \nabla_\mathbf{x} f(\mathbf{x}).
\end{equation*}
Clearly, this gradient indicates how changes in the input ($\mathbf{x}$) affect the output of $f$. As we are interested in the Expected Gradient, we average the gradients over a distribution $p(\bar{\mathbf{x}})$ of baseline inputs $\bar{\mathbf{x}}$, namely:
\begin{equation*}
\underset{\mathbf{\bar{\mathbf{x}}} \sim p(\bar{\mathbf{x}})}{\mathbb{E}} [ \nabla_\mathbf{x} f(\bar{\mathbf{x}})].    
\end{equation*}
Finally, the feature-attribution value $\phi^i$ for each feature $i$ is computed by multiplying the difference between the input $\mathbf{x}$ and the baseline $\bar{\mathbf{x}}$ with the expected gradient:
\begin{align*}
\phi^{i} = \underset{\mathbf{\bar{\mathbf{x}}} \sim p(\bar{\mathbf{x}})}{\mathbb{E}} \bigg[ \left(x^i - \bar{x}^i\right) \frac{\partial f(\mathbf{\bar{\mathbf{x}}})}{\partial{x^i}}\bigg].
\end{align*}
EG is considered a robust feature-attribution method as it averages multiple baselines drawn from a distribution $p(\bar{\mathbf{x}})$.

\section{Methodology}

The egRUE framework comprises two components:
\begin{itemize}
    \item \textbf{egRUE}: An improved uncertainty estimate informed by XAI feature attribution scores.
    \item \textbf{egRUE Explanations}: The decomposition of egRUE to derive feature-wise uncertainty contributions.
\end{itemize}

We start by introducing \emph{uncertainty estimation}.
\subsection{Uncertainty Estimation}
Uncertainty estimation methods aim to quantify the level of uncertainty associated with a model's predictions.
\begin{definition}[Uncertainty Estimation]\label{def:uncertainty}
    Given an instance $\mathbf{x} \in \mathbb{R}^J$, an {\em uncertainty estimation} method is a function $\sigma : \mathbb{R}^J \rightarrow \mathbb{R}$ that takes $\mathbf{x}$ and quantifies uncertainty with a real value for some prediction function applied to $\mathbf{x}$.
\end{definition}

\subsubsection{RUE}
In this paper, we build on the Reconstruction Uncertainty Estimate (RUE), an AutoEncoder-inspired uncertainty estimation method \cite{LiRong1,korte2024confidence}. We start by giving a formal definition of RUE.

Consider a dataset $D = \{(\mathbf{x}_i, \mathbf{y}_i) \mid i=1, \ldots, n\}$, each instance is composed of an input vector ($\mathbf{x}\in \mathbb{R}^J$) and an output vector ($\mathbf{y}\in \mathbb{R}^o$). By minimising prediction errors on the training set, we derive a prediction model.
The prediction model can often be decomposed into two components: 
\begin{enumerate}
\item The encoder $f_{\phi} : \mathbb{R}^J \rightarrow \mathbb{R}^k$;
\item The predictor head $f_{\psi} : \mathbb{R}^k \rightarrow \mathbb{R}^o$;
\end{enumerate}
The encoder encodes the input into $k$ salient features in the latent space while the predictor uses the $k$ features from the encoder to predict the output. 

In addition to the encoder and the predictor, RUE trains an additional component, 
the decoder ($g: \mathbb{R}^k \rightarrow \mathbb{R}^J$). This decoder 
reconstructs the input $\mathbf{x}$ given the $k$ encoded salient features from the encoder, minimising the reconstruction errors in the training set.
With the fit decoder, RUE is defined as the reconstruction error of the decoder for input $\mathbf{x}$.

Formally, the reconstruction error in RUE serves as the uncertainty estimate. The higher the reconstruction error, the greater the uncertainty in the prediction.
\begin{definition}[RUE Uncertainty]
The {\em RUE Uncertainty} for an instance $\mathbf{x}$ and its reconstruction $\hat{\mathbf{x}}$ is:\footnote{For a vector $v$, $\|v\|_1$ denotes the L1 norm of $v$.} 
    \begin{align*}
        \sigma_{R}(\mathbf{x}) =  \| \mathbf{x} - \hat{\mathbf{x}} \|_1.
    \end{align*}
\end{definition}
\noindent
Intuitively, RUE uncertainty holds as $\sigma_{R}$ is expected to show concordance 
with prediction error:
\begin{equation*}
   \|\mathbf{y}-f_{\psi} \circ f_{\phi}(\mathbf{x})\|_1  \leftrightarrow \|\mathbf{x}-g \circ f_{\phi}(\mathbf{x})\|_1.
\label{eqn:rue_n_pe}
\end{equation*}
This assertion was motivated by two assumptions: 
\begin{enumerate}
\item Predictor performs well on instances that are similar to the training set and poorly on dissimilar instances.
\item The decoder which has been trained on the same training set, would reconstruct samples similar to the training set better, yielding a low reconstruction error and vice versa. 
\end{enumerate}

\subsubsection{egRUE}
In RUE, uncertainty is computed as the sum of feature-wise reconstruction errors, assuming that each feature contributes equally to prediction uncertainty. However, we hypothesize that features do not contribute evenly to uncertainty, just as they do not contribute evenly to the prediction. Important features contribute more to uncertainty, as the model's inability to reconstruct them indicates higher uncertainty.

Hence, we propose a second uncertainty estimation method, egRUE (Expected Gradients RUE), which extends RUE with feature attribution values from XAI. Before we can compute egRUE, we formally define feature-wise reconstruction errors: 

\begin{definition}[Feature-wise Reconstruction Error]\label{defn:fw_re}
The feature-wise reconstruction error $\reconstructionError_{\mathbf{x}}$ between an instance $\mathbf{x}  = \langle x^1, \ldots, x^J\rangle$ and its associated reconstructed instance $\hat{\mathbf{x}} = \langle \hat{x}^1, \ldots, \hat{x}^J\rangle$ is:
\begin{equation*}
    \reconstructionError_{\mathbf{x}} = \langle |x^1 - \hat{x}^1|, \ldots, |x^J - \hat{x}^J|\rangle.
\end{equation*}
\end{definition}

In this work, we adapt EG for feature-attribution computation in egRUE with the following modifications:
(1) using the encoder and predictor head $f_{\psi} \circ f_{\phi}$ of RUE as the prediction model; 
(2) changing the baseline distribution to sampled instances from the training data distribution $\mathcal{D}$; and
(3) using the input instance as the target instance. 
Thus, the feature-attribution value for each feature $j$ is:
\begin{align*}
    &\phi^{j}_\mathbf{x} = \underset{\mathbf{x}_i \sim \mathcal{D}}{\mathbb{E}} \bigg[\left(x^j - x_i^j\right) \frac{\partial (f_{\psi} \circ f_{\phi}(\mathbf{x}_i))}{\partial x^j}\bigg]. 
\end{align*}

egRUE adjusts the contribution of each feature to prediction uncertainty by weighting feature-wise reconstruction errors with reference to their feature attribution values. These feature-importance-weights are then calculated by normalizing the absolute EG feature-attribution values.
\begin{align*}
    \featureWeights{j} = \frac{\left|\phi^{j}_\mathbf{x}\right|}{\sum_{j=1}^J{\left|\phi^{j}_\mathbf{x}\right|}}.
\end{align*}
We denote the feature-importance weights as a vector: 
\begin{align*}
    \omega_{\mathbf{x}} = \langle \featureWeights{1},
    \ldots,
    \featureWeights{j}\rangle,
\end{align*}
and define egRUE as the feature-importance-weighted sum of feature-wise reconstruction errors. Formally,
\begin{definition}[egRUE Uncertainty]
Given an instance $\mathbf{x}$, let $\reconstructionError_{\mathbf{x}}$ and $\omega_{\mathbf{x}}$ be its feature-wise reconstruction error and feature-importance weights, respectively. The {\em egRUE Uncertainty} of $\mathbf{x}$ is $\sigma_{e}(\mathbf{x})$:
\begin{align*}
    \sigma_{e}(\mathbf{x}) =  \reconstructionError_{\mathbf{x}} \cdot \omega_{\mathbf{x}}.
\end{align*}
\end{definition}
\noindent
Compared to RUE uncertainty ($\sigma_{R}$), egRUE uncertainty ($\sigma_{e}$) provides a more \enquote{targeted} estimation by explicitly considering feature-attribution values computed using EG, using the same prediction model.

\subsection{Uncertainty Explanation}
To explain uncertainty estimation, we again consider feature attribution. For an uncertainty estimation $\sigma$, we want to see how does each feature contribute to $\sigma$. 
\begin{definition}[Uncertainty Explanation] \label{uncertainty_ft_att}
Given an instance $\mathbf{x}$, an {\em uncertainty explanation method} $\zeta$ 
is a mapping $\zeta(\mathbf{x}): \mathbb{R}^J \rightarrow \mathbb{R}^J$, with 
each $\zeta^i$ quantifying the prediction uncertainty contributed by the corresponding $x^i$. 
\end{definition}

To compute explanations for RUE and egRUE, we observe that both $\sigma_{R}$ and $\sigma_{e}$ are aggregate values of feature-wise reconstruction errors, discounted by prediction feature-attribution values in the case of egRUE. Thus, the feature-wise reconstruction error naturally \enquote{explains} how each feature affects the uncertainty estimation. 

For RUE, we define {\em RUE Explanation} as follows.
\begin{definition}[RUE Explanation]
Given an instance $\mathbf{x}$, its {\em RUE Explanation} $\zeta_{R}$ is its feature-wise reconstruction error.
\begin{equation*}
    \zeta_{R}(\mathbf{x}) = \reconstructionError_{\mathbf{x}}.
\end{equation*}
\end{definition}

For egRUE, 
we define egRUE explanation as follows.
\begin{definition}[egRUE Explanation]
Given an instance $\mathbf{x}$, its {\em egRUE Explanation} $\zeta_{e}$ is the Hadamard product of its RUE Explanation and feature-importance values $\omega_{\mathbf{x}}$.
\begin{align*}
\zeta_{e}(\mathbf{x}) = \reconstructionError_{\mathbf{x}} \odot \omega_{\mathbf{x}} = \langle \reconstructionError^1_{\mathbf{x}}\omega_{\mathbf{x}}^1,\ldots,\reconstructionError^J_{\mathbf{x}}\omega_{\mathbf{x}}^J \rangle.
\end{align*}
\end{definition}
\noindent
Intuitively, as $\omega_{\mathbf{x}}$ is used to discount features-attributions in uncertainty estimation $\sigma_{e}$, the same discount factor is applied in the corresponding explanations in egRUE. 

\section{Experiments}
We present the results of our experiments evaluating uncertainty estimates and their explanations for egRUE. Our evaluation covers two main aspects:
\begin{enumerate}
    \item \textbf{Quality of Uncertainty Estimates}: 
    Evaluating the reliability, misclassification detection, selective prediction, robustness to false negatives, and out-of-distribution detection of these estimates.
    \item \textbf{Informativeness of Uncertainty Explanations}: Assessing whether uncertain features influence prediction error, evaluated qualitatively and quantitatively using \enquote{Remove and Retrain} (ROAR) \cite{NEURIPS2019_fe4b8556}.
\end{enumerate}

Our experiments show that egRUE consistently outperforms existing uncertainty estimation methods across reliability, misclassification detection, selective prediction, and robustness to false negatives. Relative to state-of-the-art Deep Ensembles \cite{wang2025uncertainty,zaidi2021neural,rahaman2021uncertainty,ovadia2019can}, egRUE achieves gains of 151–330\% in reliability, 4–53\% in misclassification detection, 32–77\% in selective prediction, and 78–98\% in robustness to false negatives, while improving out-of-distribution detection by 11-97\%. Moreover, egRUE provides explanations that better align with true feature-level uncertainties than RUE.

\begin{table*}[htbp]
\centering
\caption{Performance comparison of uncertainty estimates on four datasets. The best-performing uncertainty estimates for each metric have been \textbf{bolded}, and the next best have been \underline{underlined}.}
\scriptsize
\setlength{\tabcolsep}{2.5pt}
\begin{tabular}{lcccccccc}
\toprule
    \multirow{2}{*}{\textbf{UE}} & \multicolumn{4}{c}{\textbf{Lung Cancer}} & \multicolumn{4}{c}{\textbf{Colorectal Cancer}} \\
    \cmidrule(lr){2-5} \cmidrule(lr){6-9}
    & \textbf{Corr ($\uparrow$)} & \textbf{AUROC ($\uparrow$)} & \textbf{AURC ($\downarrow$)}  & \textbf{$\sigma$-Risk ($\downarrow$)} &\textbf{Correlation ($\uparrow$)} & \textbf{AUROC ($\uparrow$)} & \textbf{AURC ($\downarrow$)} & \textbf{$\sigma$-Risk ($\downarrow$)}\\
\midrule
    egRUE & \textbf{0.688 $\pm$ 0.151} & \textbf{0.835 $\pm$ 0.101} & \underline{0.015 $\pm$ 0.006} & \underline{0.008 $\pm$ 0.001} & \underline{0.602 $\pm$ 0.047} & \textbf{0.812 $\pm$ 0.013} & 0.044 $\pm$ 0.015 & \underline{0.016 $\pm$ 0.004} \\
\midrule
    Entropy & \underline{0.676 $\pm$ 0.121} & 0.814 $\pm$ 0.078 & 0.024 $\pm$ 0.023 & 0.026 $\pm$ 0.014 & \textbf{0.703 $\pm$ 0.035} & \textbf{0.812 $\pm$ 0.019} & 0.045 $\pm$ 0.020 & \textbf{0.000 $\pm$ 0.000} \\
    MCD & 0.300 $\pm$ 0.244 & 0.665 $\pm$ 0.133 & 0.039 $\pm$ 0.034 & \textbf{0.000 $\pm$ 0.000} & -0.531 $\pm$ 0.301 & 0.307 $\pm$ 0.217 & 0.222 $\pm$ 0.113 & 0.441 $\pm$ 0.324 \\
    DE & 0.179 $\pm$ 0.671 & 0.617 $\pm$ 0.320 & 0.064 $\pm$ 0.088 & 0.065 $\pm$ 0.094 & -0.262 $\pm$ 0.314 & 0.531 $\pm$ 0.240 & 0.121 $\pm$ 0.105 & 0.164 $\pm$ 0.217 \\
    PN Alea & 0.486 $\pm$ 0.112 & \underline{0.830 $\pm$ 0.066} & 0.036 $\pm$ 0.025 & 0.010 $\pm$ 0.002 & 0.386 $\pm$ 0.141 & \underline{0.741 $\pm$ 0.137} & \underline{0.031 $\pm$ 0.012} & 0.020 $\pm$ 0.004 \\
    PN Epis & 0.184 $\pm$ 0.136 & 0.725 $\pm$ 0.113 & 0.053 $\pm$ 0.035 & 0.109 $\pm$ 0.077 & 0.122 $\pm$ 0.091 & 0.653 $\pm$ 0.147 & 0.043 $\pm$ 0.026 & 0.071 $\pm$ 0.061 \\
    GPC & 0.115 $\pm$ 0.000 & 0.702 $\pm$ 0.000 & \textbf{0.010 $\pm$ 0.000} & 0.010 $\pm$ 0.000 & 0.067 $\pm$ 0.000 & 0.615 $\pm$ 0.000 & \textbf{0.018 $\pm$ 0.000} & 0.017 $\pm$ 0.000 \\
\midrule
\midrule
    \multirow{2}{*}{\textbf{UE}} & \multicolumn{4}{c}{\textbf{OCTMNIST}} & \multicolumn{4}{c}{\textbf{BloodMNIST}} \\
    \cmidrule(lr){2-5} \cmidrule(lr){6-9}
    & \textbf{Corr ($\uparrow$)} & \textbf{AUROC ($\uparrow$)} & \textbf{AURC ($\downarrow$)}  & \textbf{$\sigma$-Risk ($\downarrow$)} &\textbf{Correlation ($\uparrow$)} & \textbf{AUROC ($\uparrow$)} & \textbf{AURC ($\downarrow$)} & \textbf{$\sigma$-Risk ($\downarrow$)}\\

\midrule
    egRUE & \textbf{0.439 $\pm$ 0.039} & \textbf{0.786 $\pm$ 0.021} & \underline{0.120 $\pm$ 0.025} & \textbf{0.020 $\pm$ 0.016} & \textbf{0.552 $\pm$ 0.047} & \textbf{0.930 $\pm$ 0.025} & \textbf{0.139 $\pm$ 0.013} & \textbf{0.003 $\pm$ 0.003} \\
\midrule
    Entropy & -0.018 $\pm$ 0.066 & 0.681 $\pm$ 0.034 & 0.173 $\pm$ 0.043 & 0.161 $\pm$ 0.037 & -0.113 $\pm$ 0.050 & 0.668 $\pm$ 0.056 & 0.306 $\pm$ 0.067 & 0.330 $\pm$ 0.061 \\
    MCD & -0.026 $\pm$ 0.063 & 0.665 $\pm$ 0.033 & 0.176 $\pm$ 0.043 & 0.162 $\pm$ 0.041 & -0.104 $\pm$ 0.059 & 0.667 $\pm$ 0.054 & 0.307 $\pm$ 0.066 & 0.330 $\pm$ 0.060 \\
    DE & \underline{0.124 $\pm$ 0.032} & \underline{0.757 $\pm$ 0.017} & \textbf{0.113 $\pm$ 0.019} & \underline{0.089 $\pm$ 0.017} & 0.220 $\pm$ 0.143 & 0.864 $\pm$ 0.033 & 0.204 $\pm$ 0.034 & 0.157 $\pm$ 0.040 \\
    PN Alea & -0.046 $\pm$ 0.042 & 0.644 $\pm$ 0.052 & 0.260 $\pm$ 0.089 & 0.263 $\pm$ 0.076 & 0.222 $\pm$ 0.120 & \underline{0.871 $\pm$ 0.042} & 0.202 $\pm$ 0.026 & 0.222 $\pm$ 0.037 \\
    PN Epis & -0.051 $\pm$ 0.129 & 0.579 $\pm$ 0.129 & 0.296 $\pm$ 0.115 & 0.298 $\pm$ 0.129 & 0.218 $\pm$ 0.165 & 0.864 $\pm$ 0.057 & 0.205 $\pm$ 0.032 & 0.184 $\pm$ 0.045 \\
    BNN & -0.007 $\pm$ 0.022 & 0.701 $\pm$ 0.009 & 0.198 $\pm$ 0.011 & 0.176 $\pm$ 0.017 & 0.094 $\pm$ 0.055 & 0.799 $\pm$ 0.021 & \underline{0.165 $\pm$ 0.024} & \underline{0.114 $\pm$ 0.015} \\
    DEC & 0.081 $\pm$ 0.022 & 0.624 $\pm$ 0.022 & 0.259 $\pm$ 0.008 & 0.174 $\pm$ 0.033 & \underline{0.383 $\pm$ 0.023} & 0.770 $\pm$ 0.016 & 0.223 $\pm$ 0.007 & 0.490 $\pm$ 0.137 \\
\bottomrule
\end{tabular}
\label{tab:ue_perf}
\end{table*}

\begin{table}[ht]
\centering
\caption{Out-of-distribution (OOD) detection AUROC on image datasets. OOD datasets are ordered by increasing dissimilarity from the training set. Best uncertainty estimates are \textbf{bolded}; second-best are \underline{underlined}.}
\setlength{\tabcolsep}{3pt}
\scriptsize
\begin{tabular}{lcccc}
\toprule
    \multirow{2}{*}{\textbf{UE}} & \multicolumn{3}{c}{\textbf{OCTMNIST ($\uparrow$)}} \\
    \cmidrule{2-4}
    & \textbf{OCTDL-IN} & \textbf{OCTDL-OUT} & \textbf{ChestMNIST} \\
\midrule
    egRUE & \textbf{0.961 $\pm$ 0.015} & \textbf{0.977 $\pm$ 0.007} & \textbf{1.000 $\pm$ 0.000} \\
\midrule
    Entropy & 0.674 $\pm$ 0.031 & 0.386 $\pm$ 0.066 & 0.826 $\pm$ 0.060 \\
    MCD & 0.661 $\pm$ 0.034 & 0.392 $\pm$ 0.064 & 0.844 $\pm$ 0.060 \\
    DE & 0.739 $\pm$ 0.009 & 0.497 $\pm$ 0.029 & \underline{0.900 $\pm$ 0.028} \\
    PN Alea & 0.625 $\pm$ 0.063 & 0.492 $\pm$ 0.025 & 0.500 $\pm$ 0.107 \\
    PN Epis & 0.565 $\pm$ 0.052 & 0.483 $\pm$ 0.087 & 0.470 $\pm$ 0.145 \\
    BNN & \underline{0.742 $\pm$ 0.014} & \underline{0.734 $\pm$ 0.016} & 0.896 $\pm$ 0.015 \\
    DEC & 0.633 $\pm$ 0.012 & 0.669 $\pm$ 0.011 & 0.760 $\pm$ 0.013 \\
\midrule
\midrule
    \multirow{2}{*}{\textbf{UE}} & \multicolumn{3}{c}{\textbf{BloodMNIST ($\uparrow$)}} \\
    \cmidrule{2-4}
    & \textbf{Raabin-IN} & \textbf{BoneMarrow-OUT} & \textbf{ChestMNIST} \\
\midrule
    egRUE & \textbf{0.997 $\pm$ 0.001} & \textbf{0.946 $\pm$ 0.021} & \textbf{0.998 $\pm$ 0.001} \\
\midrule
    Entropy & 0.657 $\pm$ 0.092 & 0.884 $\pm$ 0.031 & 0.670 $\pm$ 0.221 \\
    MCD & 0.657 $\pm$ 0.090 & 0.880 $\pm$ 0.028 & 0.683 $\pm$ 0.215 \\
    DE & 0.847 $\pm$ 0.050 & \underline{0.943 $\pm$ 0.008} & 0.798 $\pm$ 0.106 \\
    PN Alea & \underline{0.879 $\pm$ 0.058} & 0.882 $\pm$ 0.016 & 0.905 $\pm$ 0.048 \\
    PN Epis & 0.875 $\pm$ 0.070 & 0.871 $\pm$ 0.028 & 0.912 $\pm$ 0.070 \\
    BNN & 0.820 $\pm$ 0.026 & 0.874 $\pm$ 0.014 & \underline{0.914 $\pm$ 0.015} \\
    DEC & 0.764 $\pm$ 0.018 & 0.808 $\pm$ 0.022 & 0.897 $\pm$ 0.037 \\
\bottomrule
\end{tabular}
\label{tab:ood_detection_perf}
\end{table}

\subsection{Datasets}
We conducted experiments on four medical diagnosis datasets spanning various domains, modalities, and sizes.

\paragraph{Singapore Chinese Health Study Lung Cancer}
This dataset was derived from the Singapore Chinese Health Study (SCHS) \cite{hankin2001singapore}, which investigates the influence of diet, genetics, and environment on cancer and other chronic illnesses in the ethnic Chinese population in Singapore. The dataset comprises 23,622 instances. Each instance represents a participant and is characterized by 20 features describing their diet, genetic makeup, and environment, along with a target label indicating whether the participant suffers from lung cancer. After preprocessing, 17,464 data instances remained, of which 1.855\% were lung cancer cases.

\paragraph{Singapore Chinese Health Study Colorectal Cancer}
This SCHS-derived dataset \cite{hankin2001singapore} contains 22,901 instances and shares features with the Lung Cancer dataset, except for sleep duration and includes colorectal cancer polygenic scores. After preprocessing, 20,689 instances remained, with 2.576\% labelled as colorectal cancer.

\paragraph{OCTMNIST}
This dataset \cite{kermany2018identifying} contains 109,309 retinal OCT images (224×224), split into training (97,477), validation (10,832), and test (1,000) sets. Class distribution is imbalanced: 47\% Normal, 34\% Choroidal Neovascularization (CNV), 11\% Diabetic Macular Edema (DME), and 8\% Drusen. To assess uncertainty under distributional shift, we augment the test set with 618 OCTDL \cite{kulyabin2024octdl} instances from overlapping classes. For OOD detection, we use: (1) OCTDL-IN (same classes), (2) OCTDL-OUT (unseen OCTDL classes, e.g., AMD), and (3) 1,000 ChestMNIST X-ray images \cite{wang2017chestx}.

\paragraph{BloodMNIST}
This dataset \cite{acevedo2020dataset} contains 17,092 blood cell microscopic images (224×224), split into training (11,959), validation (1,712), and test (3,421). It is imbalanced, with 19.5\% of instances from the platelet class in the 8-class task. To assess performance under distributional shift, we augment the test set with 3,421 Raabin \cite{kouzehkanan2021raabin} instances. For OOD detection, we use: (1) Raabin (overlapping classes), (2) 3,421 bone marrow (from unseen classes) \cite{matek2021highly}, and (3) 3,421 ChestMNIST X-ray images \cite{wang2017chestx}.

\subsection{Uncertainty Estimation Baselines}
We compare egRUE with standard uncertainty baselines, including Entropy \cite{malininPredictiveUncertaintyEstimation2018}, Bayesian Neural Networks (BNN) \cite{mackay1992practical}, Monte Carlo Dropout (MCD) \cite{galDropoutBayesianApproximation2016}, Deep Ensembles (DE) \cite{lakshminarayanan2017simple}, Deep Evidential Classification (DEC) \cite{sensoyEvidentialDeepLearning2018a}, Posterior Networks (PN) \cite{charpentierPosteriorNetworkUncertainty2020}, and Gaussian Process (GP) \cite{williams1995gaussian}.

\begin{figure*}[th]
\centering
\begin{subfigure}[t]{0.49\textwidth} 
    \centering
    \begin{subfigure}[t]{\explImageProp\linewidth}
       \centering
       \includegraphics[width=\linewidth]{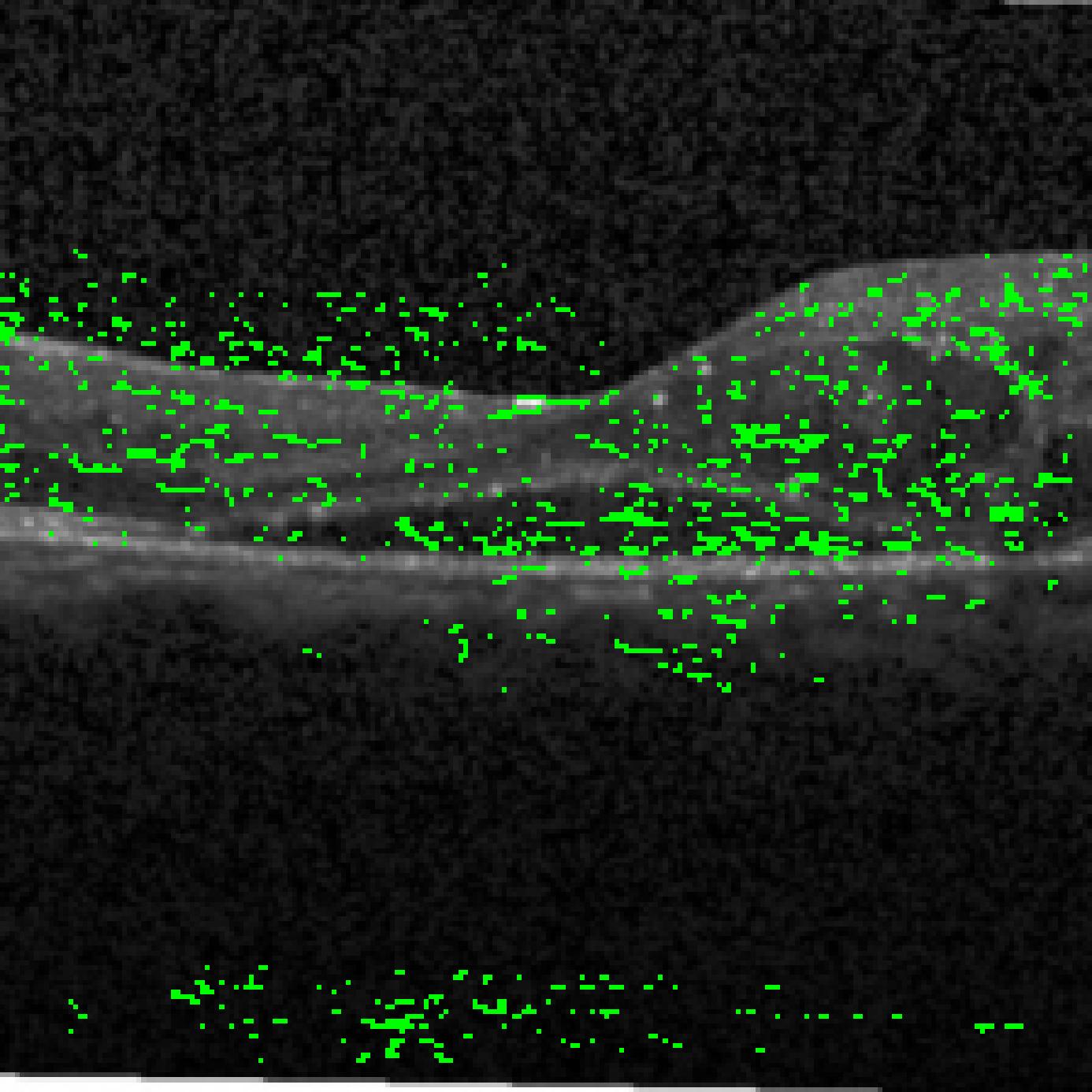}
       \caption*{EG}
    \end{subfigure}
    \begin{subfigure}[t]{\explImageProp\linewidth}
       \centering
       \includegraphics[width=\linewidth]{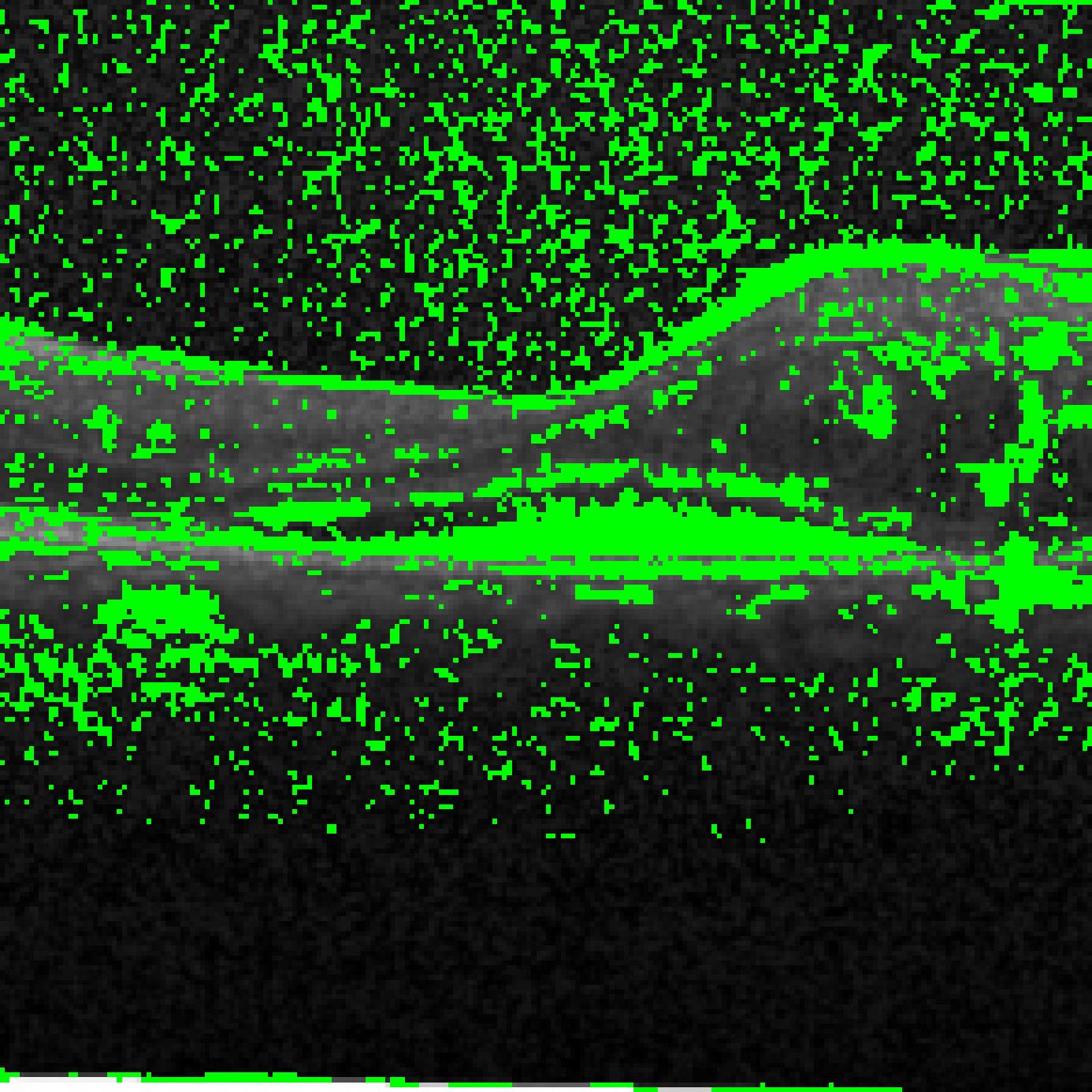}
       \caption*{RUE}
    \end{subfigure}
    \begin{subfigure}[t]{\explImageProp\linewidth}
       \centering
       \includegraphics[width=\linewidth]{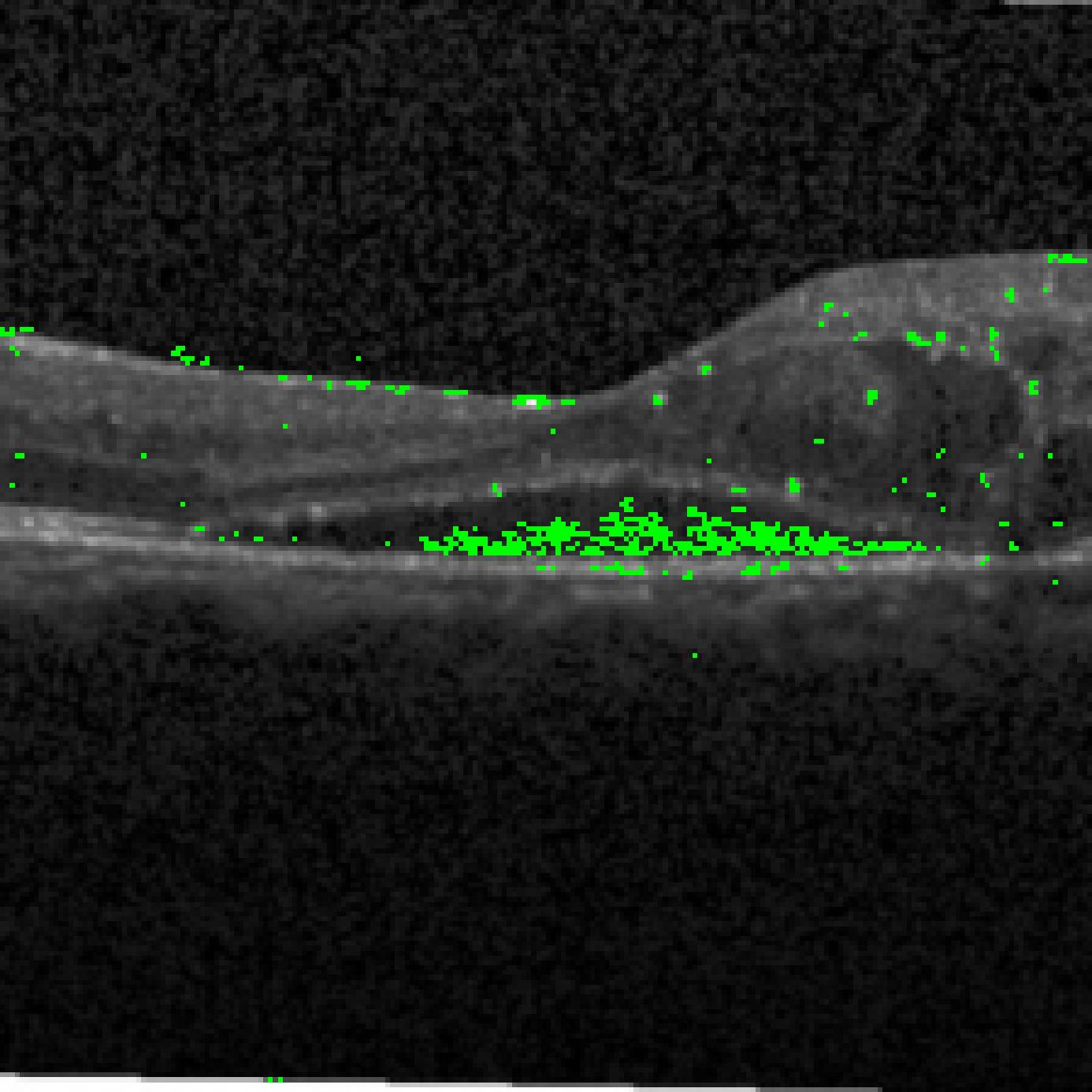}
       \caption*{egRUE}
    \end{subfigure}
    \caption{OCTMNIST}
    \label{fig:expl_images_oct}
\end{subfigure}
\begin{subfigure}[t]{0.49\textwidth} 
    \centering
    \begin{subfigure}[t]{\explImageProp\linewidth}
       \centering
       \includegraphics[width=\linewidth]{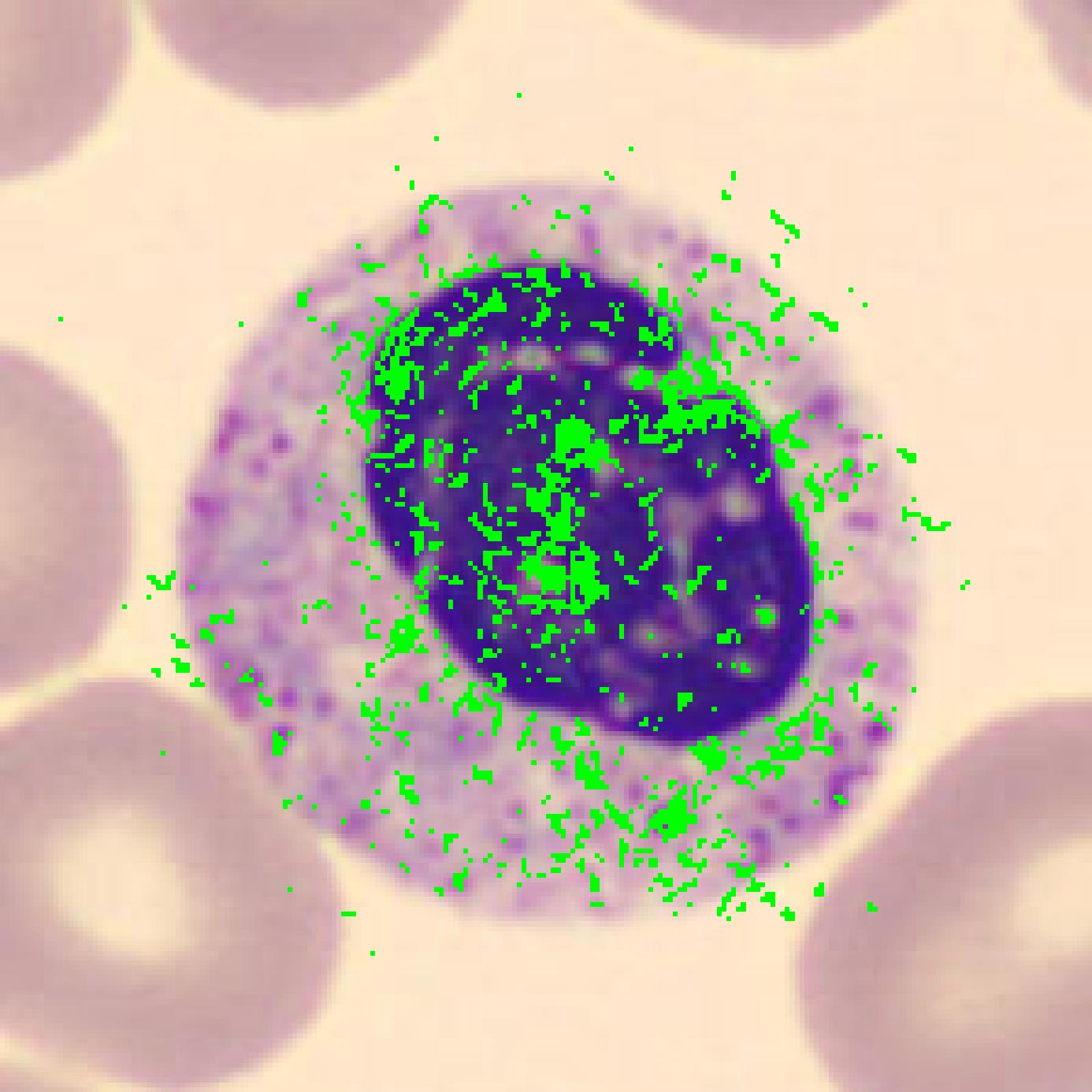}
       \caption*{EG}
    \end{subfigure}
    \begin{subfigure}[t]{\explImageProp\linewidth}
       \centering
       \includegraphics[width=\linewidth]{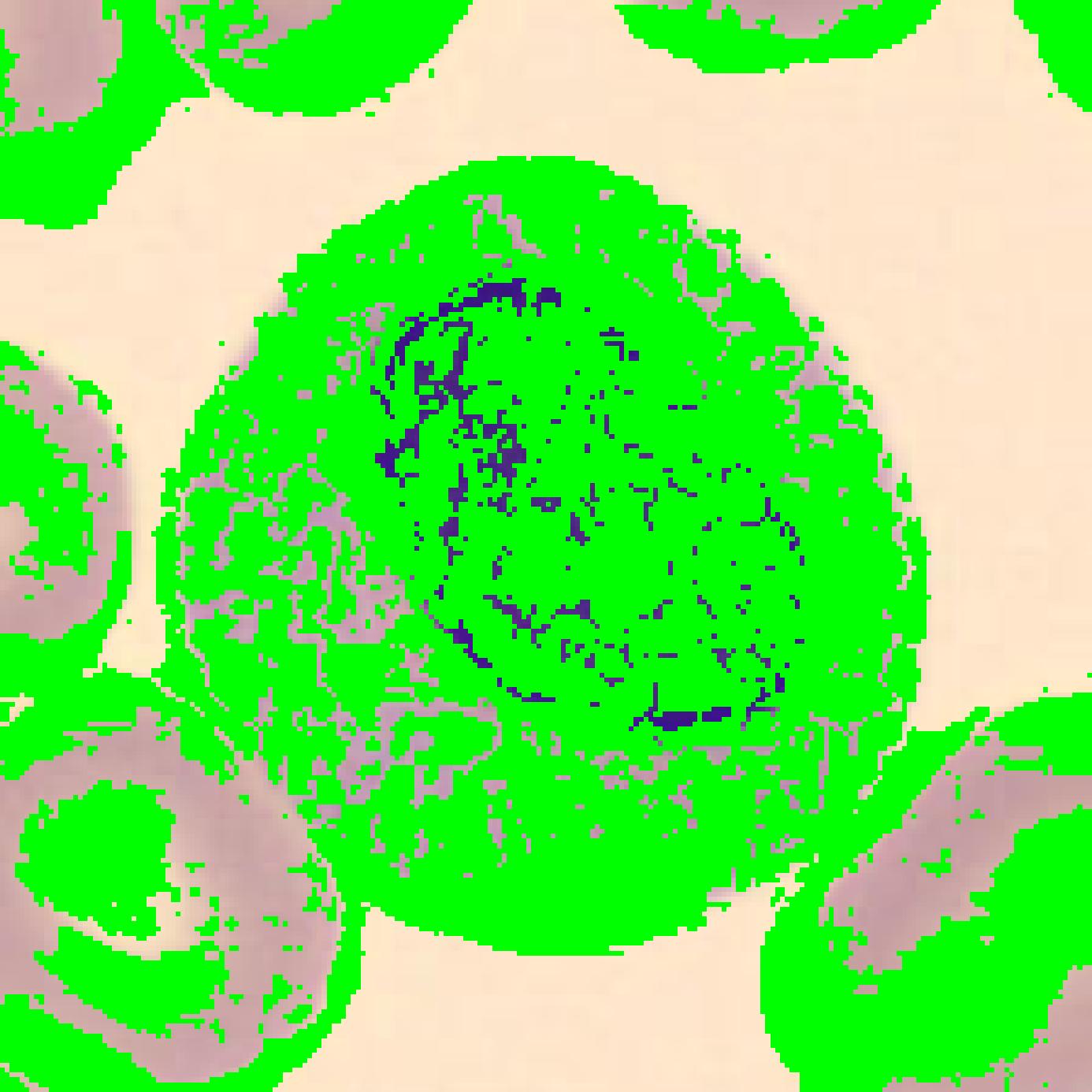}
       \caption*{RUE}
    \end{subfigure}
    \begin{subfigure}[t]{\explImageProp\linewidth}
       \centering
       \includegraphics[width=\linewidth]{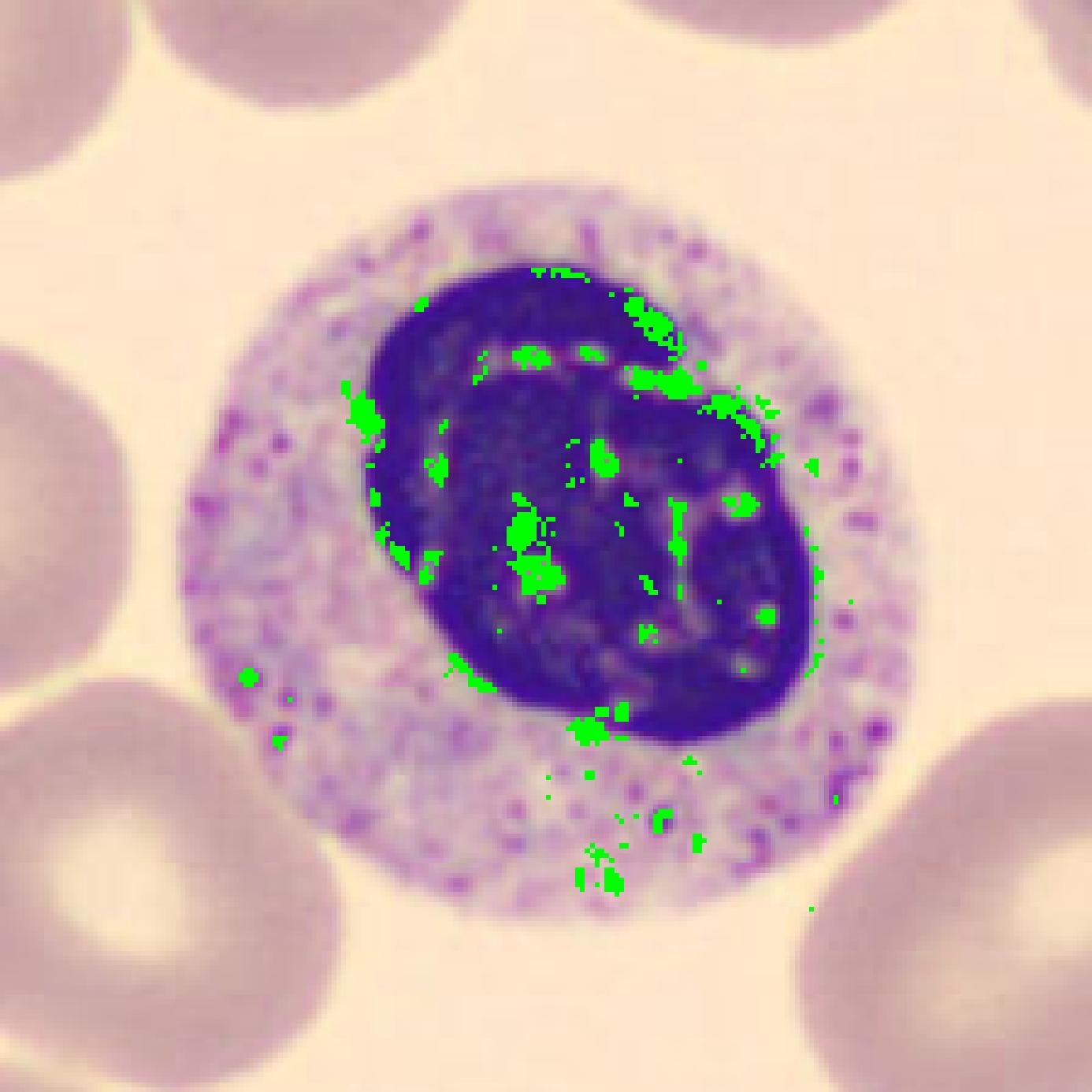}
       \caption*{egRUE}
    \end{subfigure}
    \caption{BloodMNIST}
    \label{fig:expl_images_blood}
\end{subfigure}
\caption{Comparison of Expected Gradients (EG), RUE, and egRUE for mispredicted \enquote{Diabetic Macular Edema (DME)} (OCTMNIST) and \enquote{Immature Granulocytes} (BloodMNIST) images. Green highlights mark regions driving incorrect predictions (EG) or predictive uncertainty (RUE, egRUE). 
(\ref{fig:expl_images_oct}) EG highlights most of the retina and some artefacts without pinpointing any specific region of uncertainty. RUE significantly overestimates uncertainty regions with even more artefacts. egRUE highlights the edema (triangular region) typical of DME cases. (\ref{fig:expl_images_blood}) EG highlights most of the central cell, RUE highlights all cells in the image, while egRUE focuses on the nucleus, whose shape is atypical of a neutrophil, driving model uncertainty.
}
\label{fig:expl_images}
\vspace{-10pt}
\end{figure*}

\subsection{Metrics}
Five metrics are used for evaluation in this work.
\paragraph{Correlation}\cite{miTrainingFreeUncertaintyEstimation2022,upadhyayBayesCapBayesianIdentity2022} Quantifies the reliability of the uncertainty estimate using the Pearson correlation between the cross-entropy loss and the uncertainty estimate. A high correlation indicates a reliable uncertainty estimate. 
\paragraph{Misclassification Prediction with AUROC} \cite{corbiere2019addressing,zhu2022rethinking} Measures how well uncertainty scores distinguish correct from incorrect predictions, using 0/1 loss as ground truth and uncertainty as prediction scores.
\paragraph{AURC} \cite{dingRevisitingEvaluationUncertainty2020} The Area Under the Risk-Coverage Curve measures an uncertainty estimate's effectiveness for selective prediction, where risk is 0/1 loss and coverage is the proportion of certain predictions.
\paragraph{$\sigma$-Risk Score} \cite{wang2025towards} Measures average 0/1 loss on confident instances (instances with normalized uncertainty below $\sigma=0.1$). Low $\sigma$-risk indicates robustness to false negatives (i.e., confident but incorrect predictions).
\paragraph{OOD Detection Performance} \cite{lakshminarayanan2017simple,malininPredictiveUncertaintyEstimation2018} Measures how well uncertainty distinguishes in-distribution from OOD instances using AUROC, with dataset labels (0=in, 1=OOD) as ground truth and uncertainty as prediction scores.

\subsection{Quantitative Evaluation}
\subsubsection{Uncertainty Estimation}
Table~\ref{tab:ue_perf} evaluates uncertainty estimates in terms of reliability (correlation), selectivity (AUROC, AURC), and robustness to false negatives ($\sigma$-risk). We observe that egRUE consistently performs well across all three aspects and all four datasets. It shows strong reliability, achieving the highest correlation with error on the Lung Cancer, OCTMNIST and BloodMNIST datasets, and the second highest on the Colorectal dataset. In terms of selectivity, egRUE yields the highest AUROC for misclassification prediction on across all datasets, and achieves the lowest and second lowest AURC on Lung Cancer, OCTMNIST, and BloodMNIST datasets -- demonstrating its effectiveness for selective prediction. Finally, examining $\sigma$-Risk scores, egRUE shows strong robustness to false negatives, achieving either the lowest or second lowest error on confident predictions across all datasets.

Table~\ref{tab:ood_detection_perf} compares the OOD detection performance of uncertainty estimates on the image datasets. egRUE consistently achieves the highest AUROC on both OCTMNIST and BloodMNIST, across OOD datasets with varying degrees of dissimilarity to the training set, indicating superior OOD detection performance. 

\subsubsection{Uncertainty Explanation}
Figure~\ref{fig:expl_images} compares explanations from EG, RUE, and egRUE. In~\ref{fig:expl_images_oct}, EG and RUE produce noisy explanations, whereas egRUE clearly highlights the edema-associated triangular region \cite{panozzo2004diabetic}. In~\ref{fig:expl_images_blood}, EG produces noisy explanations, and RUE indiscriminately highlights all cells, offering little clinically specific insight. In contrast, egRUE focuses on clinically relevant features: it highlights the indented, rather than uniformly curved, nucleus shape which is atypical of a mature neutrophil but typical of an immature granulocyte. Both examples demonstrate egRUE’s superior ability to identify clinically meaningful uncertain features.

\begin{figure}[ht]
\centering 
    \includegraphics[width=0.9\linewidth]{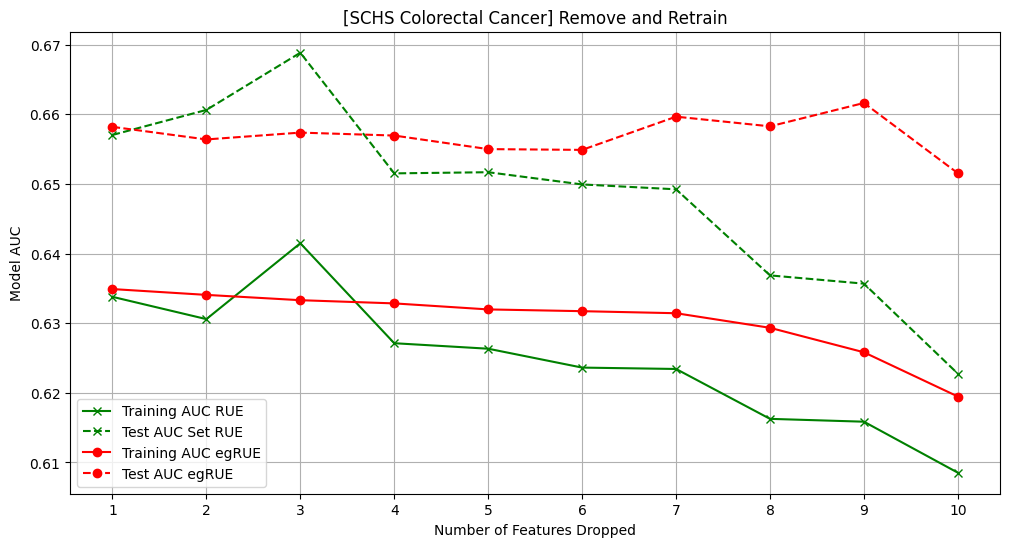}
    \caption{Results of ROAR experiments on the Colorectal Cancer dataset. egRUE yields a more stable, higher AUC than RUE after removing uncertain features, indicating its uncertainty explanations better reflect true uncertainties.}
    \label{fig:SCHS_ROAR}
    \vspace{-15pt}
\end{figure}

Figure~\ref{fig:SCHS_ROAR} evaluates RUE and egRUE uncertainty explanations with the Remove and Retrain (ROAR) metric \cite{NEURIPS2019_fe4b8556}. This approach assesses the effects of removing uncertain features and retraining the prediction model. Due to class imbalances in all datasets, we used AUC (Area Under the receiver operating characteristic Curve) instead of accuracy for ROAR. The results show that egRUE generally yields a more stable and higher AUC, indicating that egRUE explanations better align with each feature's true uncertainty. 

\subsubsection{Ablation: egRUE vs RUE}
Comparing egRUE and RUE estimates (Table~\ref{tab:ue_perf_ablation}), we find that egRUE outperforms RUE across all datasets, highlighting the benefits of incorporating feature-attribution weights into RUE’s uncertainty estimation.
\begin{table}[ht]
\centering
\scriptsize
\caption{Performance comparison of egRUE and RUE on four datasets. The best-performing uncertainty estimates for each metric have been \textbf{bolded}.}
\setlength{\tabcolsep}{2.5pt}
\begin{tabular}{p{1cm}lcccc}
\toprule
    \cmidrule(lr){2-5} 
    \textbf{Data} & \textbf{UE} &\textbf{Corr ($\uparrow$)} & \textbf{AUROC ($\uparrow$)} & \textbf{AURC ($\downarrow$)} & \textbf{$\sigma$-Risk ($\downarrow$)}\\
\midrule
    \multirow{2}{1cm}{Lung Cancer} & egRUE & \textbf{0.688 $\pm$ 0.151} & \textbf{0.835 $\pm$ 0.101} & \textbf{0.015 $\pm$ 0.006} & \textbf{0.008 $\pm$ 0.001} \\
    & RUE & 0.589 $\pm$ 0.115 & 0.796 $\pm$ 0.092 & 0.021 $\pm$ 0.011 & 0.019 $\pm$ 0.003 \\
\midrule
    \multirow{2}{1cm}{Colorectal Cancer} & egRUE & \textbf{0.602 $\pm$ 0.047} & \textbf{0.812 $\pm$ 0.013} & \textbf{0.044 $\pm$ 0.015} & \textbf{0.016 $\pm$ 0.004} \\
    & RUE & 0.542 $\pm$ 0.059 & 0.794 $\pm$ 0.026 & 0.049 $\pm$ 0.021 & 0.018 $\pm$ 0.003 \\
\midrule
    \multirow{2}{1cm}{OCT MNIST} & egRUE & \textbf{0.439 $\pm$ 0.039} & \textbf{0.786 $\pm$ 0.021} & \textbf{0.120 $\pm$ 0.025} & \textbf{0.020 $\pm$ 0.016} \\
    & RUE & 0.395 $\pm$ 0.059 & 0.771 $\pm$ 0.034 & 0.129 $\pm$ 0.028 & 0.051 $\pm$ 0.047 \\
\midrule
    \multirow{2}{1cm}{Blood MNIST} & egRUE & 0.552 $\pm$ 0.047 & \textbf{0.930 $\pm$ 0.025} & \textbf{0.139 $\pm$ 0.013} & \textbf{0.003 $\pm$ 0.003} \\
    & RUE & \textbf{0.583 $\pm$ 0.051} & 0.923 $\pm$ 0.024 & 0.142 $\pm$ 0.013 & 0.004 $\pm$ 0.003 \\
\bottomrule
\end{tabular}
\label{tab:ue_perf_ablation}
\end{table}

\begin{table}[ht]
\centering
\caption{Area under the ROAD curve ($\uparrow$) for each attribution method, paired with RUE.}
\setlength{\tabcolsep}{3pt}
\scriptsize
\begin{tabular}{lcC{1cm}C{1.8cm}C{2.5cm}}
\toprule
    & \textbf{egRUE} & \textbf{igRUE} & \textbf{gradCamRUE} & \textbf{guidedBackpropRUE} \\ 
\midrule
    BloodMNIST & 0.148 & 0.144 & 0.133 & 0.138 \\ 
\bottomrule
\end{tabular}
\end{table}

\begin{table}[htbp]
\centering
\setlength{\tabcolsep}{3pt}
\scriptsize
\caption{Instance inference time (s) of uncertainty estimates.}
\begin{tabular}{lccccccc}
\toprule
    ~ & \textbf{Entropy} & \textbf{DEC} & \textbf{egRUE} & \textbf{PN} & \textbf{DE} & \textbf{MCD} & \textbf{BNN} \\
\midrule
    Time / s & 0.00132 & 0.00139 & 0.00200 & 0.00407 & 0.01026 & 0.01249 & 0.01259 \\ 
\bottomrule
\end{tabular}
\label{tab:runtime}
\end{table}

\subsubsection{Ablation: Attribution Method}
We evaluate alternative attribution methods using ROAD (Remove and Debias) \cite{rong2022consistent}, which ranks features by uncertainty, progressively removes them (10–90\%), and tracks accuracy. The area under accuracy–removal curve quantifies explanation quality, with larger areas indicating stronger attribution. On BloodMNIST, egRUE achieves highest ROAD scores, indicating that alternative attribution methods do not improve uncertainty explanations.

\subsubsection{Computational Efficiency}
We compare the instance-level inference time of egRUE with several sampling-based and deterministic uncertainty methods (Table~\ref{tab:runtime}). egRUE achieves inference in 0.002s per instance, substantially faster than sampling-based methods such as DE (0.0103s), MCD (0.0125s), and BNN (0.0126s). Deterministic methods like Entropy (0.0013s) and DEC (0.0014s) are slightly faster, but egRUE provides a favorable trade-off between efficiency and robust uncertainty estimation.

\section{User Study}
To evaluate the effectiveness of our uncertainty explanations at fostering calibrated trust in addition to uncertainty scores, we conducted a user study with five medical experts on a blood cell classification task with examples from BloodMNIST. The survey consisted of nine questions evenly split across three categories to observe the impact of uncertainty explanations on calibrated trust in each category. The three categories were characterised by the model prediction’s correctness and uncertainty score: (1) correct and low uncertainty, (2) incorrect and high uncertainty, (3) incorrect and low uncertainty. The experts were first shown the blood cell image, the model’s prediction, and the uncertainty score, and then they were asked how much they agreed with the model’s prediction. Next, the experts were shown the uncertainty explanation and asked if it had influenced their confidence in the model’s prediction (increased, no change, or decreased), and to justify their change or non-change.

\begin{figure}[t]
    \centering
    \includegraphics[width=0.9\linewidth, trim=0.6cm 0.6cm 0cm 0cm, clip]{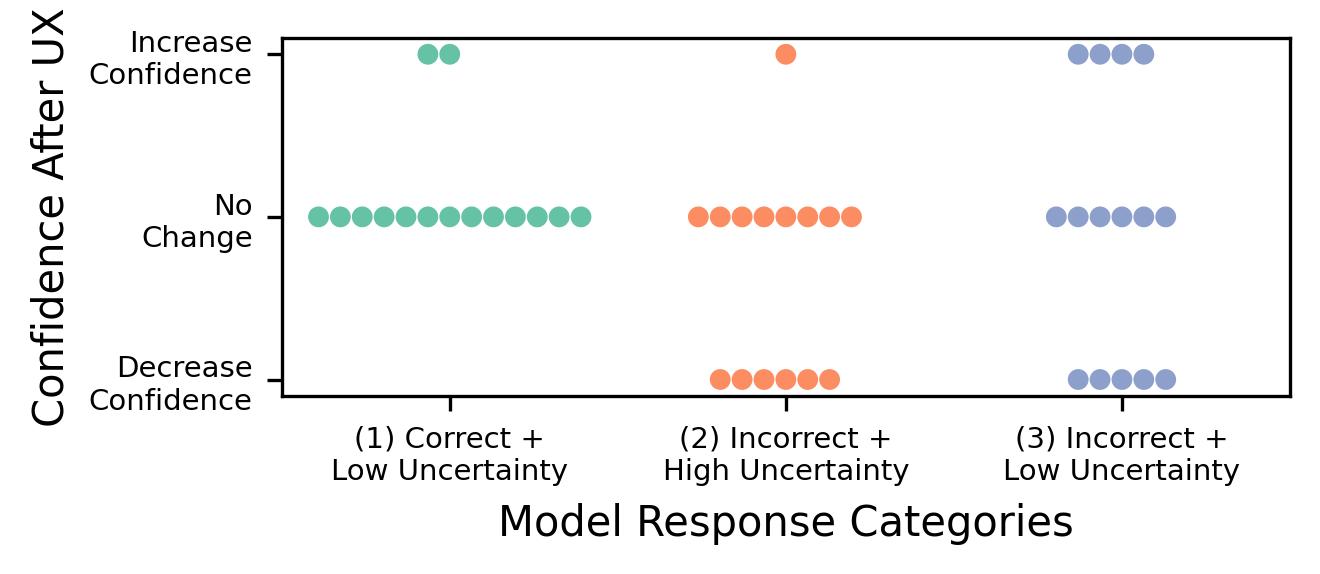}
    \caption{User Study Results: Change in respondents' confidence after the addition of uncertainty explanations.}
    \label{fig:user_study}
\end{figure}

Firstly, our user study observed that in the presence of uncertainty scores, experts are still prone to overreliance on model predictions: experts agreed with the model’s prediction 40\% (12 out of 30 responses) of the time, even though the model’s prediction was incorrect. This demonstrates that uncertainty estimates alone are not sufficient to combat overconfidence in model predictions. Observing the change in respondents’ confidence after adding explanations, we see an improvement in calibrated trust for Cases (1) and (2) (Figure~\ref{fig:user_study}), with increased confidence in correct predictions and reduced confidence in incorrect predictions. For Case (3), the effect on confidence appears mixed, but analysing the qualitative justifications for increased confidence paints a fuller picture. In all four instances, respondents indicated that their increased confidence stemmed from the uncertainty explanations being consistent with their own sources of uncertainty. 

\section{Theoretical Evaluation}
We present a theoretical analysis of RUE and egRUE, covering both uncertainty estimation and explanation. First, we evaluate how uncertainty estimates provide bounds on prediction error. Next, we prove that the egRUE feature attribution satisfies key feature-attribution properties. 
This analysis highlights the robustness and reliability of our methods in providing meaningful insights into model predictions. 

\subsection{Uncertainty Estimation}
In evaluating an uncertainty estimation method, obtaining theoretical results that bound the prediction error with uncertainty estimation is critical. Such bounding ensures that the model's confidence in its predictions aligns with its actual performance. This alignment is fundamental to the reliability of the developed uncertainty estimation. Both RUE and egRUE provide such bounds. We present the result for RUE first.\footnote{Full proofs of all theoretical results can be found in the appendix.}
\begin{theorem}\label{thrm:rue_bounds}
There exists an $\ell\in \mathbb{R}^+$ and an $r\in \mathbb{R}^+$ derived from the given prediction model, such that errors of the predictor are bounded by RUE uncertainty:
\begin{equation*}
    \ell~\sigma_{R}(\mathbf{x}) \leq \|\mathbf{y}-f_{\psi} \circ f_{\phi}(\mathbf{x})\|_1 \leq r~\sigma_{R}(\mathbf{x}).
\end{equation*}
\end{theorem}
Theorem~\ref{thrm:rue_bounds} states that there exist positive constants $\ell$ and $r$ such that for any instance $\mathbf{x}$, the difference between the model's prediction $f_{\psi} \circ f_{\phi}(\mathbf{x})$ and the ground truth $\mathbf{y}$ is greater than or equal to $\ell~\sigma_{R}(\mathbf{x})$ but less than or equal to $r~\sigma_{R}(\mathbf{x})$. Broadly speaking, this theorem holds because the prediction model and the uncertainty estimation model share the encoder $f_{\phi}$. As the prediction error is \enquote{propagated} within the two overlapping models, the error in the reconstruction infers the error in the prediction (under our assumptions of invertible layers and a well-trained prediction model).

We can prove that the egRUE counterpart to Theorem~\ref{thrm:rue_bounds} by extending the earlier proof using the following relation between RUE and egRUE:
\begin{proposition}\label{prop:egrue_lb_ub}
    For any instance $\mathbf{x}$, it holds that:
\begin{equation*}
    \sigma_e(\mathbf{x}) \leq \sigma_{R} (\mathbf{x}) \leq \frac{1}{\min_{j}{\featureWeights{j}}}~\sigma_{e}(\mathbf{x}).
\end{equation*}
\end{proposition} 

Proposition~\ref{prop:egrue_lb_ub} states that RUE is lower-bounded by egRUE and upper-bounded by egRUE multiplied by a constant. To demonstrate this, we first observe that RUE is computed as the sum of feature-wise reconstruction errors, while egRUE is the weighted sum of these errors. Additionally, we note that in egRUE, all weights are normalized to be less than 1, whereas in the upper bound, the constant ensures that the weights are greater than or equal to 1. By applying the transitive property of inequality, we can extend Theorem~\ref{thrm:rue_bounds} with Proposition~\ref{prop:egrue_lb_ub}, thereby proving Theorem~\ref{thrm:egrue_bounds}.

\begin{theorem}\label{thrm:egrue_bounds}
There exists an $\ell_e\in \mathbb{R}^+$ and an $r_e\in \mathbb{R}^+$, derived from the given prediction model, such that errors of the predictor are bounded by egRUE:
\begin{equation*}
    \ell_e~\sigma_{e}(\mathbf{x}) \leq \|\mathbf{y}-f_{\psi} \circ f_{\phi}(\mathbf{x})\|_1 \leq r_e~\sigma_{e}(\mathbf{x}).
\end{equation*}
\end{theorem}
Theorem~\ref{thrm:egrue_bounds} states that there exist positive constants $\ell_e, r_e \in \mathbb{R}^+$ such that, for any instance $\mathbf{x}$, the difference between the model's prediction $f_{\psi} \circ f_{\phi}(\mathbf{x})$ and the ground truth $\mathbf{y}$ is lower bounded by $\ell_e~\sigma_{e}(\mathbf{x})$ and upper bounded by $r_e~\sigma_{e}(\mathbf{x})$.

\subsection{Uncertainty Explanation}
Our analysis of uncertainty explanation is centered around the following three properties: (1) {\em implementation invariance}, (2) {\em sensitivity}, and (3) {\em consistency}. 
Implementation invariance states that a uncertainty feature-attribution algorithm should produce the same uncertainty feature-attribution values for functionally equivalent networks.
\begin{property}[Implementation Invariance]
Let $f$ and $f^\prime$ be two functionally equivalent prediction models.
An uncertainty explanation algorithm $\zeta$ satisfies {\em implementation invariance} if and only if for all $\mathbf{x}$, $\zeta(\mathbf{x})$ produces the same uncertainty explanation when applied to both $f$ and $f^\prime$. 
\end{property}

Two prediction models are functionally equivalent if for all the same set of inputs, both models produce the same outputs.  

The next property is {\em sensitivity}, stated as follows.
\begin{property}[Sensitivity]
If a prediction function $f$ does not depend on some feature $x^j \in \mathbf{x}$, then that feature should be assigned an uncertainty feature-attribution of zero.\footnote{A function is said to \enquote{depend} on a feature, if a change in that feature value results in a change in the function's output.} 
\end{property}
\noindent
Sensitivity is desirable because it ensures that only features that are relevant to the prediction  have non-zero uncertainty feature-attribution values assigned. 

The last property {\em consistency} states that, for two models, such that one model receives greater contribution from a feature as seen in the prediction model, then the feature cannot receive a decreased uncertainty feature-attribution value. 
\begin{property}[Consistency]
 Let $\mathbf{x}^{\backslash j}$ denote setting $x^j = 0$, then given two prediction models $f$ and $f^\prime$, an uncertainty explanation algorithm is consistent if and only if for
    \begin{align*}
    \vert f(\mathbf{x}) - f(\mathbf{x}^{\backslash j}) \vert &\geq \vert f^\prime(\mathbf{x}) - f^\prime(\mathbf{x}^{\backslash j}) \vert, ~\text{it holds that}\\
    \zeta^j(\mathbf{x}) \mbox{ for } f(\cdot) &\geq \zeta^j(\mathbf{x}) \mbox{ for } f^\prime(\cdot).  
\end{align*}
\end{property}
\noindent
Consistency is desirable as it ensures that feature attributions align with the relative importance of features across different models. This property maintains that features contributing more to the output of a model, with the same feature-wise reconstruction error, should consistently receive higher attribution values.
We first show that $\zeta_R$ satisfies the {\em implementation invariance} property. 
\begin{theorem}\label{thrm:rue_sat}
    $\zeta_R$ satisfies implementation invariance.
\end{theorem}
This is true, since the learned encoder, decoder and predictor will present the same feature-wise reconstruction error and predicted values, and thus $\zeta_R$ will inherently satisfy implementation invariance.

We proceed to show that $\zeta_e$ extends the theoretical guarantees of $\zeta_R$, satisfying the {\em implementation invariance}, {\em sensitivity} and {\em consistency} properties. 
\begin{theorem}\label{thrm:egrue_sat}
    $\zeta_e$ satisfies implementation invariance, sensitivity, and consistency.
\end{theorem}

It is clear that $\zeta_e$ follows a similar proof to Theorem~\ref{thrm:rue_sat} to satisfy implementation invariance, and as the computation of feature-attribution evaluates the gradients of the predictor with respect to the input (chain-rule), the partial derivative of the irrelevant feature will naturally present a zero-value, and therefore return a zero-value uncertainty feature-attribution score for any input feature that is not relevant to the predictor. Finally, we prove consistency, to achieve this we assert that only the prediction head $f_\psi$ is modified to evaluate consistency, thus maintaining the same reconstruction error (as we are evaluating the same instance $\mathbf{x}$), it is then a natural consequence of EG that consistency holds.

\section{Conclusion}
We propose egRUE (Expected Gradient-weighted Reconstruction Uncertainty Estimate), an uncertainty estimation method that quantifies prediction uncertainty while explaining how individual features contribute to it. By integrating Expected Gradient scores into the Reconstruction Uncertainty Estimate (RUE), egRUE provides informative and interpretable uncertainty estimates, addressing the lack of trust that limits AI adoption in medicine.

We establish egRUE's reliability through theoretical analysis and experiments on four real-world healthcare datasets. We prove egRUE bounds prediction error and demonstrate consistent improvements in misclassification detection, selective prediction, robustness to false negatives, and out-of-distribution detection compared to state-of-the-art methods. Theoretical analysis shows that egRUE explanations satisfy invariance, sensitivity, and consistency, while Remove-and-Retrain experiments confirm they align more closely with true sources of uncertainty than RUE. A user study with medical experts shows that egRUE’s explanations improve calibrated trust over uncertainty scores alone, increasing confidence in correct predictions and reducing confidence in incorrect ones. By revealing both when and why a model is uncertain, egRUE enhances trust and supports risk-aware clinical decision-making.

\section*{Acknowledgment}
This research is supported by the Singapore Ministry of Health’s National Medical Research Council under its Population Health Research Grant MOH-002116-00 and the Centre of AI in Medicine (C-AIM).

\bibliographystyle{IEEEtran}
\bibliography{citations}

\appendix
\subsection{Implementation Details}
Tabular experiments were repeated 10 times, image experiments 5 times; we report mean and standard deviation. Architecture, hyperparameters, seed settings, and computing infrastructure information are in the code repository: \href{https://github.com/lr98769/egrue\_icdm}{https://github.com/lr98769/egrue\_icdm}



\subsection{Proof of Theorem \ref{thrm:rue_bounds}}
\begin{theorem}
There exists an $\ell\in \mathbb{R}^+$ and $r\in \mathbb{R}^+$ derived from the given prediction model such that errors of the predictor are bounded by RUE uncertainty:
\begin{equation*}
    \ell~\sigma_{R}(\mathbf{x}) \leq \|\mathbf{y}-f_{\psi} \circ f_{\phi}(\mathbf{x})\|_1 \leq r~\sigma_{R}(\mathbf{x}).
\end{equation*}
\end{theorem}
To prove Theorem~\ref{thrm:rue_bounds}, we assume the predictor model is a multi-layer perceptron (MLP) with $L$ layers, each containing $k$ nodes and employing a linear activation function. Each layer comprises a weight matrix $W^{[l]}$ and a bias vector $b^{[l]}$. Layers 1 to $M$ of the predictor model are referred to as the encoder, while layers $M+1$ to $L$ constitute the predictor head. The forward propagation operation of each layer is described as:
\begin{equation}\label{eqn:fwd_prop}
Z^{[l]} = W^{[l]}Z^{[l-1]}+b^{[l]}.
\end{equation}

We define \enquote{Layer-wise Error} ($\Delta^{[l]}$) as a vector which fulfills the following constraint:
\begin{equation}\label{eqn:lw_error}
Z^{[l]} + \Delta^{[l]} = W^{[l]}\left(Z^{[l-1]} + \Delta^{[l-1]}\right)+b^{[l]}.
\end{equation}

Using Equation~\ref{eqn:fwd_prop} and \ref{eqn:lw_error}, we can simplify the definition of \enquote{Layer-wise Error} ($\Delta^{[l]}$) to:
\begin{equation}\label{eqn:lw_error_fwd}
    \Delta^{[l]} = W^{[l]}\Delta^{[l-1]}.
\end{equation}

Following Equation~\ref{eqn:lw_error_fwd} and assuming that  $W^{[l]}$ is invertible, we arrive at another definition of \enquote{Layer-wise Error} ($\Delta^{[l]}$):
\begin{equation}\label{eqn:lw_error_bwd}
    \Delta^{[l-1]} = {W^{[l]}}^{-1}\Delta^{[l]}.
\end{equation}

Finally, we define the \enquote{Layer-wise Loss} ($\mathcal{L}^{[l]}$) of layer $l$ as the L1 norm of \enquote{Layer-wise Error} in layer $l$:
\begin{equation}\label{eqn:lw_loss}
    \mathcal{L}^{[l]} = \|\Delta^{[l]}\|_1=\sum_{i=1}^k\left|\delta_i^{[l]}\right|.
\end{equation}
\noindent where $\delta_i^{[l]}$ represents the $i$-th element of $\Delta^{[l]}$.

We decompose Theorem~\ref{thrm:rue_bounds} into 2 bounds relating layer-wise loss at layer $M$ with prediction and reconstruction error: 
\begin{alignat*}{2}
\exists~\ell_p, r_p,\ell_d, r_d\in &\mathbb{R}^+ s.t. \\
    \ell_p~\mathcal{L}^{[M]} &\leq \|\mathbf{y}-f_{\psi} \circ f_{\phi}(\mathbf{x})\|_1 &&\leq r_p~\mathcal{L}^{[M]}, \\
    \ell_d~\sigma_{R}(\mathbf{x}) &\leq \mathcal{L}^{[M]} &&\leq r_d~\sigma_{R}(\mathbf{x}).
\end{alignat*}

Since both the decoder and predictor are MLPs, we can generalise both bounds to the following lemma:
\begin{lemma}\label{lemma:mlp_bounds}
In an MLP with $L$ layers, where each layer $l$ has a layer-wise loss $\mathcal{L}^{[l]}$, there exists $p,q\in\mathbb{R}^{+}$ such that: 
\begin{align}
    \mathcal{L}^{[L]} &\leq p \times \mathcal{L}^{[l]} \label{eqn:lemma_lb}, \\
    \mathcal{L}^{[l]} &\leq q \times \mathcal{L}^{[L]} \label{eqn:lemma_ub}, \text{for}~l\in \{1, \ldots, L\}.
\end{align}
\end{lemma}

To demonstrate the first statement (Equation~\ref{eqn:lemma_lb}) in Lemma~\ref{lemma:mlp_bounds}, we make the following observations:
\begin{enumerate} 
    \item The layer-wise loss $\mathcal{L}^{[l]}$ in layer $l$ is bounded from above by maximum per-output loss $\left|\delta_i^{[l]}\right|$ multiplied by $k$.
\begin{equation} \label{eqn:rue_obs1}    
    \mathcal{L}^{[l]} =\sum_{i=1}^k\left|\delta_i^{[l]}\right| \leq k\times \max_{i\in\{1,\ldots, k\}}{\left|\delta_i^{[l]}\right|}.
\end{equation} 
\item Following Equations~\ref{eqn:lw_error_fwd} and~\ref{eqn:lw_loss} and applying the triangle inequality, the per-output loss is bounded from above by the maximum weight in each node $\left|W_{ij}^{[l]}\right|$ multiplied by the layer-wise loss from the previous layer $\mathcal{L}^{[l-1]}$.
\begin{equation} \label{eqn:rue_obs2}
    \left|\delta_i^{[l]}\right| \leq \max_{j\in\{1,\ldots, k\}}{\left|W_{ij}^{[l]}\right|} \times \mathcal{L}^{[l-1]}.
\end{equation}
\end{enumerate}
By applying the above two observations recursively with the transitive property of inequalities, we obtain the following inequality (Equation~\ref{eqn:rue_lb_proved}), which shows that there exists a $p$ such that Equation~\ref{eqn:lemma_lb} is satisfied, thereby proving the first statement of Lemma~\ref{lemma:mlp_bounds}. 
\begin{equation}\label{eqn:rue_lb_proved}
    \mathcal{L}^{[L]}\leq \underbrace{k^{L-l} \times \prod_{d\in\{l+1, \ldots, L\}}{\max_{i,j\in\{1, \ldots, k\}}{\left|W_{ij}^{[d]}\right|}}}_{p}\times \mathcal{L}^{[l]}.
\end{equation}

The second statement of Lemma~\ref{lemma:mlp_bounds} (Equation~\ref{eqn:lemma_ub}) can be proved similarly. First, we derive the following inequality using the triangle inequality and Equations~\ref{eqn:lw_error_bwd} and~\ref{eqn:lw_loss}.
\begin{equation} \label{eqn:rue_obs3}
    \left|\delta_i^{[l]}\right| \leq \max_{j\in\{1,\ldots, k\}}{\left|{W^{[l+1]}}_{ij}^{-1}\right|} \times \mathcal{L}^{[l+1]}.
\end{equation}

Applying Equation~\ref{eqn:rue_obs1} and Equation~\ref{eqn:rue_obs3} recursively with the transitive property of inequalities, we obtain the following inequality (Equation~\ref{eqn:rue_ub_proved}), demonstrating that there exists a $q$ such that Equation~\ref{eqn:lemma_ub} is satisfied, thereby proving the second statement of Lemma~\ref{lemma:mlp_bounds}. 
\begin{equation} \label{eqn:rue_ub_proved}
    \mathcal{L}^{[l]}\leq \underbrace{k^{L-l} \times \prod_{d\in\{l+1, \ldots, L\}}{\max_{i,j\in\{1, \ldots, k\}}{\left|{{W^{[d]}}}^{-1}_{ij}\right|}}}_q\times \mathcal{L}^{[L]}.
\end{equation}

With both statements of Lemma~\ref{lemma:mlp_bounds} proven, we have demonstrated that Theorem~\ref{thrm:rue_bounds} holds. 

\subsection{Proof of Theorem \ref{thrm:rue_sat}}
To prove implementation invariance for $\zeta_R$, we note that by definition one assumes the same input ($\mathbf{x}$) and the same output ($g$, $f_{\psi} \circ f_{\phi}(\mathbf{x})$) of any two functionally equivalent networks. Therefore, since $\zeta_R$ is simply the returned feature-wise reconstruction errors given as an output from the egRUE network architecture, regardless of implementation, the reconstruction error will be maintained for the same input and outputs. 

\subsection{Proof of Theorem \ref{thrm:egrue_sat}}
From Theorem~\ref{thrm:rue_sat}, it is clear that the feature-wise reconstruction error in $\zeta_R$ satisfies implementation invariance. Thus, it is enough to show that EG satisfies the implementation invariance property in order for implementation invariance to hold for $\zeta_e$. As presented by the authors in \cite{IG} (Remark 3), EG is a reformulation of IG to an expectation over many baselines, and any feature-attribtuion method of the same form satisfies implementation invariance. 

To show this, we evaluate the gradients used for the explanation, which naturally satisfy implementation invariance, as one can simply look at the chain-rule for the gradients. For egRUE, one can look at the input $\mathbf{x}$, encoder $f_\phi$ and predictor head $f_\psi$, where $f = f_{\psi} \circ f_{\phi}$. From this, we note that: 
\begin{align*}
    \frac{\partial f}{\partial x^j} = \frac{\partial f_{\psi} \circ f_{\phi}}{\partial x^j} = \frac{\partial f_\psi}{\partial f_\phi} \cdot \frac{\partial f_\phi}{\partial x^j},~\text{for all $j$}
\end{align*}
Thus, the intermediary layers of the network are the implementation details described by $f_{\phi}$. Since one can ignore the $f_{\phi}$ term for any two functions, by back-propagation, it is clear the explanation given by the gradient of any two functions will be the same, assuming the same input and output.

Following this is the sensitivity property, it is clear the reliance on the gradients from the predictor head ensures sensitivity holds with respect to the input, as if any $x^j$ is irrelevant to $f_\psi$ the returned value will be zero. Therefore, the $\zeta^j_e$ explanation is given such that: 
\begin{align*}
    \zeta^j_{e} &= |x^j - \hat{x}^j| \cdot \frac{\left|0\right|}{\sum_{j=1}^J{\left|\phi^{j}_\mathbf{x}\right|}} = |x^j - \hat{x}^j| \cdot 0 = 0.
\end{align*}
Consistency follows naturally, if one considers the same input instance and thus same $\zeta_R$, the change in prediction probability is reflected by EG, and therefore $\zeta_e$ is the normalised reflection of this change. To achieve this, we can consider two prediction heads, $f_\psi$ and $f^\prime_\psi$, such that we have the following $f^\prime = f_\phi \circ f^\prime_\psi$, to show consistency holds, we can simply consider a single baseline instance $\bar{\mathbf{x}} \sim p(\bar{\mathbf{x}})$ and input $\mathbf{x}$, and rewrite the expectation as an integral over an single instance. We can simply note that,
\begin{align*}
    f(\mathbf{x}) - f(\bar{\mathbf{x}}) =  \int^{1}_{\alpha=0} \nabla_{\mathbf{x}} f(\gamma(\alpha)) \cdot \frac{d \gamma}{d \alpha} d \alpha,~\text{and}, 
\end{align*}
\begin{align*}
    f^\prime(\mathbf{x}) - f^\prime(\bar{\mathbf{x}}) =  \int^{1}_{\alpha=0} \nabla_{\mathbf{x}} f^\prime(\gamma(\alpha)) \cdot \frac{d \gamma}{d \alpha} d \alpha.
\end{align*}
where $\gamma(\alpha) = \langle \gamma^1(\alpha), \ldots, \gamma^J(\alpha) \rangle$ is the parameterization of $\mathbf{x}$ at a point $\alpha \in [0,1]$ along a path from a baseline instance $\bar{\mathbf{x}}$ to the instance to explain $\mathbf{x}$.

The gradient $\nabla_\mathbf{x} f$ (resp. $\nabla_\mathbf{x} f^\prime$) contains the partial derivatives of each $x^j$ with respect to $f$ (resp. $f^\prime$), and since each partial derivative reflects the importance of each $x^j$ with respect to $f$ (resp. $f^\prime$), we can see a change in the predictor $f$ with respect to an input feature $x^j$ is the same regardless of implementation detail. Thus, it is reasonable to assess the magnitude of $\phi^j_{\mathbf{x}}$ ($\vert \phi^j_{\mathbf{x}} \vert$). By identifying this change in $f$ (resp. $f^\prime$) with respect to a single $x^j$, consistency holds. 

\end{document}

%% file: constants.tex
\newcommand{\paperTitle}{Explainable Uncertainty Estimation for Reliable Medical AI}

\newcommand{\explImageProp}{0.28}